\documentclass[letterpaper, 10 pt, conference]{ieeeconf}  

\IEEEoverridecommandlockouts                              

\usepackage{algpseudocode}
\usepackage{subcaption}
\usepackage{float}
\usepackage{graphicx}
\usepackage{booktabs}
\usepackage{adjustbox}
\usepackage{amsmath}
\usepackage{multirow}
\usepackage{makecell}
\usepackage[table,xcdraw]{xcolor}
\usepackage{amsmath,amssymb}
\usepackage{bm}
\usepackage[linesnumbered,ruled,vlined]{algorithm2e}

\title{\LARGE \bf
ATA: Bridging Implicit Reasoning with Attention-Guided and Action-Guided Inference for Vision-Language Action Models
}

\author{
        Cheng Yang$^{1*}$, Jianhao Jiao$^{2*}$, Lingyi Huang$^{1}$, Jinqi Xiao$^{1}$, Zhexiang Tang$^{1}$, Yu Gong$^{1}$, \\
        Yibiao Ying$^{1}$, Yang Sui$^{3}$, Jintian Lin$^{4}$, Wen Huang$^{4}$, Bo Yuan$^{1\dagger}$%
\thanks{$^{*}$Equal Contribution.}%
\thanks{$^{\dagger}$Corresponding Author.}%
\thanks{$^{1}$Rutgers University, $^{2}$University College London, $^{3}$Rice University,}
\thanks{$^{4}$TCL High-Tech Development Co., Ltd.}
}

\begin{document}

\maketitle
\thispagestyle{empty}
\pagestyle{empty}

\begin{abstract}
Vision-Language-Action (VLA) models rely on current observations, including images, language instructions, and robot states, to predict actions and complete tasks. While accurate visual perception is crucial for precise action prediction and execution, recent work has attempted to further improve performance by introducing explicit reasoning during inference. However, such approaches face significant limitations. They often depend on data-intensive resources such as Chain-of-Thought (CoT) style annotations to decompose tasks into step-by-step reasoning, and in many cases require additional visual grounding annotations (e.g., bounding boxes or masks) to highlight relevant image regions. Moreover, they involve time-consuming dataset construction, labeling, and retraining, which ultimately results in longer inference sequences and reduced efficiency. To address these challenges, we propose ATA, a novel training-free framework that introduces implicit reasoning into VLA inference through complementary attention-guided and action-guided strategies. Unlike CoT or explicit visual-grounding methods, ATA formulates reasoning implicitly by integrating attention maps with an action-based region of interest (RoI), thereby adaptively refining visual inputs without requiring extra training or annotations. ATA is a plug-and-play implicit reasoning approach for VLA models, lightweight yet effective. Extensive experiments show that it consistently improves task success and robustness while preserving, and even enhancing, inference efficiency.

\end{abstract}

\section{Introduction}

\label{sec:intro}

\subsection{Motivation}
Vision-Language Action (VLA) models \cite{kim2025openvla} have made significant strides in advancing the interpretation and synthesis of multimodal information, bridging the gap between textual, visual, and action-based inputs. Rather than using modular designs, VLA models establish a new paradigm for connecting multimodal information. Building on the success of large language models, this paradigm directly leverages the characteristics of large-scale data training, aiming to achieve highly generalizable and versatile robot action models. These models combine visual inputs, textual instructions, and robot states, such as an image with accompanying instructional text, to predict the required actions and execute tasks in the physical world. By processing and integrating these diverse modalities, VLA models have demonstrated strong capabilities in a variety of embodied tasks, such as object manipulation, long-horizon planning, and real-world household task execution.

\subsection{Challenges}
Despite their impressive capabilities, VLA models face several challenges that hinder their scalability and practical applicability. The former is mostly hindered by the high data collection cost, while the latter impedes the model training.

From the \textit{data collection} perspective, existing reasoning-based methods heavily rely on data-intensive resources. Tasks involving fine-grained manipulations require datasets with a high degree of precision, and explicit reasoning approaches further demand step-by-step annotations such as Chain-of-Thought (CoT) \cite{wei2022chain}. These requirements make large-scale data acquisition extremely costly and labor-intensive.

From the \textit{annotation requirement} perspective, many methods often rely on extra supervisory annotations, such as region-level labels or pixel-wise masks, to highlight relevant parts of the input. While these annotations can improve model perception, they are typically obtained either through costly manual labeling or by leveraging external models (e.g., object detectors or segmenters), which makes the process time-consuming, expensive, and difficult to scale across diverse environments and object categories.

The \textit{training} issues are also considerable. Pure VLA models, which directly map multimodal inputs to actions without additional reasoning, often remain fragile in complex tasks. To address this limitation, reasoning-based methods such as CoT have been proposed to enhance VLA performance. However, these methods themselves introduce significant drawbacks, requiring costly and time-consuming dataset construction, labeling, and retraining. Furthermore, they introduce longer inference sequences, which reduces inference efficiency. Training these large-scale VLA models, which can have billions of parameters, also demands extensive computational resources, often requiring hundreds of GPU-hours and substantial time investment \cite{kim2025openvla}.

\subsection{Contributions}
To address these challenges, we propose ATA (\textbf{AT}tention-Guided and \textbf{A}ction-Guided inference), a training-free framework that introduces implicit reasoning during inference. ATA integrates attention maps with the action region of interest (RoI) to refine visual inputs adaptively in a single forward pass. Instead of slowing down inference, ATA improves efficiency by introducing implicit reasoning, leading to shorter inference time while simultaneously enhancing model performance.

Our contributions are summarized as follows:
\begin{itemize}
    \item We propose ATA, a novel training-free framework introducing implicit reasoning into VLA models during inference.
    \item We introduce an attention-guided and action-guided approach that enhances visual input based on attention maps and action-based regions of interest (RoI).
    \item We conduct extensive experiments on multiple state-of-the-art models, including OpenVLA \cite{kim2025openvla}, $\pi0$-fast \cite{black2024pi_0} \cite{pertsch2025fast},  HybridVLA \cite{liu2025hybridvla}, and GR00T-N1.5 \cite{bjorck2025gr00t}, demonstrating the effectiveness and efficiency of ATA in improving VLA model performance across various tasks in simulation and real-world environments. Specifically, in the LIBERO \cite{liu2023libero} environment, ATA improves the performance of the OpenVLA model by 5.2\% and the $\pi_{0}$-fast model by 2.0\%. In the RLBench \cite{james2020rlbench} environment, ATA enhances the HybridVLA model by 5.3\%. In real-world block-stacking tasks using GR00T-N1.5, where the robot stacks three-layer towers with only $3\,$cm$\times3\,$cm$\times3\,$cm blocks, ATA achieves up to a 10\% performance improvement in complex scenarios.
\end{itemize}

\section{Related Work}
\label{sec:related}

\subsection{Vision-Language Action Models} 
Recent advances in VLA models have made significant strides in integrating vision, language, and action modalities, with models such as OpenVLA \cite{kim2025openvla}, OpenVLA-OFT \cite{kim2025fine}, $\pi_0$ \cite{black2024pi_0}, and $\pi_0$-fast \cite{pertsch2025fast}.
OpenVLA utilizes an auto-regressive architecture, processing visual inputs alongside language instructions to predict robot actions, demonstrating strong generalization in diverse manipulation tasks. 
$\pi_0$ \cite{black2024pi_0} enhances this paradigm by introducing flow matching to generate continuous action distributions, and $\pi_0$-fast \cite{pertsch2025fast} leverages autoregressive VLAs for highly dexterous and high-frequency tasks where standard discretization methods fail completely. Meanwhile, HybridVLA \cite{liu2025hybridvla} further integrates diffusion and autoregressive methods within a unified architecture, aiming to exploit the generative strength of diffusion and the efficiency of autoregression to achieve robust performance across long-horizon tasks. These models contribute to better action prediction and offer promising solutions for general robot control. 
However, due to the use of Vision-Language Models (VLMs) ~\cite{karamcheti2024prismatic}~\cite{wang2024qwen2}~\cite{team2024gemma}, these VLA models require billion-level parameters, making them computationally expensive even when techniques like Low-Rank Adaptation (LoRA) \cite{hu2022lora} are used for fine-tuning. This makes them reliant on substantial computational resources, which poses challenges in scalability and efficiency, especially in resource-constrained environments. This motivates the need for lightweight, training-free approaches that can enhance VLA performance during inference without retraining or additional annotations.

\subsection{Reasoning for Vision-Language Action Models} 
To address the fragility of VLAs in complex tasks, recent works have incorporated explicit reasoning to improve robustness and interpretability. CoT-VLA \cite{zhao2025cot} introduces visual CoT reasoning, where tasks are decomposed into intermediate reasoning steps to guide decision-making. While this improves interpretability, it requires frame-level alignment across visual, language, and action modalities. Such alignment demands highly precise annotations for each step, which makes dataset collection extremely costly and time-consuming, and inference becomes inefficient due to longer reasoning sequences. 
Contrastive Region Guidance (CRG) \cite{wan2024contrastive} improves grounding by refining attention with region-based contrastive objectives. CRG relies on external multimodal models to explicitly mark foreground objects of interest, while suppressing or masking out other objects. This explicit region annotation enhances robustness but requires heavy dependence on external annotations and models, making data preparation labor-intensive and less scalable to broader domains. From Seeing to Doing bridges symbolic reasoning and visuomotor control by combining high-level planning with low-level execution policies. While this improves transparency, the approach requires computationally expensive training and leads to slower inference, thus limiting its applicability in large-scale or real-time VLA scenarios. API \cite{yu2024attention} exploits attention maps from the Vision Transformer (ViT) \cite{dosovitskiy2020image} during inference to enhance image inputs, thereby improving the performance of Vision-Language Models (VLMs). While API shares a similar spirit of leveraging implicit reasoning at inference time, it relies solely on ViT attention maps that are agnostic to the action modality, limiting its applicability to VLA models. In contrast, ATA further introduces an action-guided strategy that encodes end-effector motion intent into directional RoIs, along with a frequency-controlled mechanism to balance performance and efficiency. Furthermore, MOKA \cite{liu2024moka} proposes mark-based visual prompting, where affordance reasoning is transformed into visual question-answering by annotating marks on images. 
Although this method enables effective zero-shot generalization, it still relies on manual design of prompts and marks, which constrains scalability. 
Finally, OneTwoVLA \cite{lin2025onetwovla} unifies reasoning and acting by adaptively switching between explicit reasoning and direct action generation, but still depends on large-scale reasoning-centric data and suffers from inference latency. Overall, existing reasoning-based methods improve interpretability and long-horizon planning, yet often face trade-offs between performance, data cost, and efficiency. In contrast, ATA is a training-free framework that injects implicit reasoning into VLA inference via attention- and action-guided strategies, improving both performance and efficiency without retraining or additional supervision.

\section{Methodology}
\label{sec:method}


In this section, we introduce the ATA inference framework for Visual-Language-Action (VLA) models, which integrates both Attention-guided and Action-guided strategies. First, we define the overall problem setup in Section \ref{sec:problem}. We then present a detailed breakdown of the Attention-guided strategy in Section \ref{sec:attention} and the Action-guided strategy in Section \ref{sec:action}. Finally, we describe how these two strategies are integrated during the inference process in Section \ref{sec:inference} to optimize model performance.

\begin{figure*}[t]
  \includegraphics[width=1\linewidth]{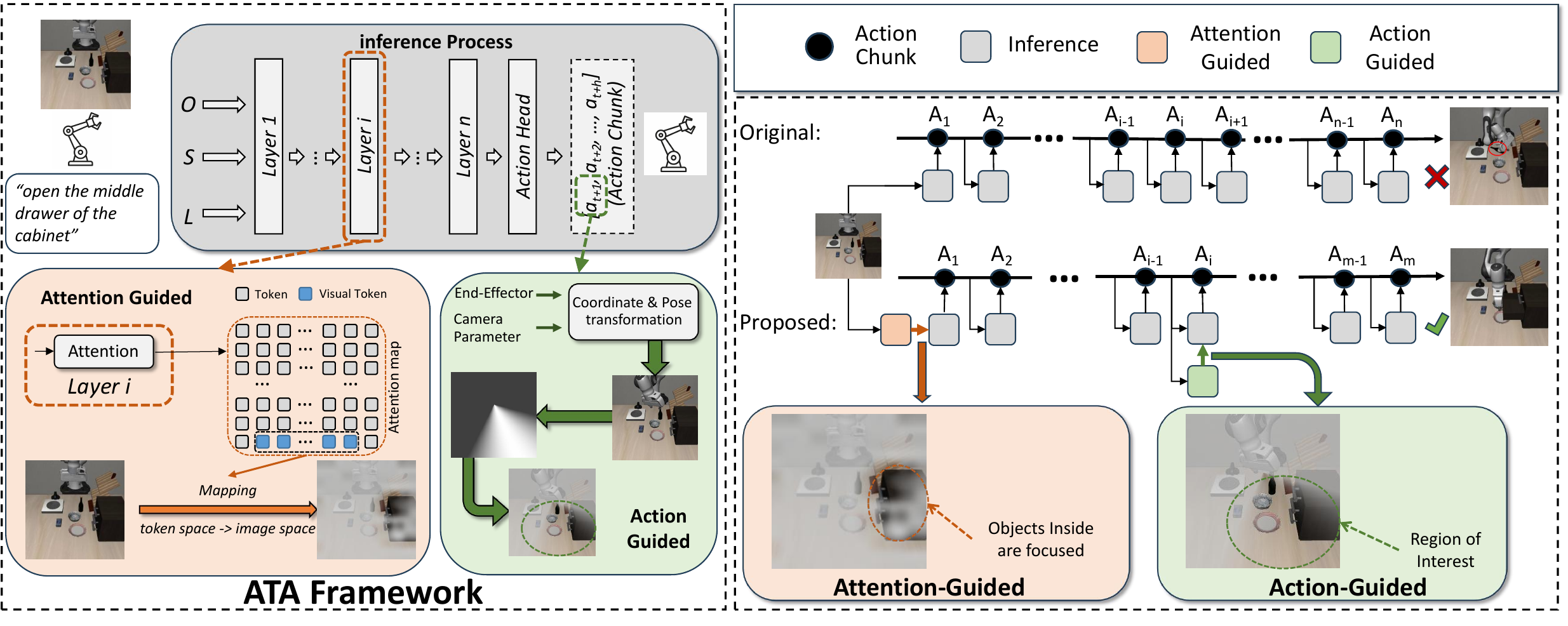}
\caption{Overview of the proposed ATA framework. (\textbf{Left}) Given a language instruction and multimodal observations, the VLA model processes tokens through stacked layers to produce an action chunk. 
Our framework injects implicit reasoning cues via two complementary strategies: (\textit{i}) Attention-guided reasoning extracts attention maps from intermediate layers to focus on task-relevant visual regions; (\textit{ii}) Action-guided reasoning leverages the end-effector pose and camera parameters to construct a directional region of interest (RoI). (\textbf{Right}) In the original pipeline (top), an error at an early step may propagate along the horizon (up to $n$ steps), leading to task failure (marked with \textcolor{red}{$\times$}) and risky behaviors such as knocking over a bottle (highlighted with a \textcolor{red}{red circle}). The proposed pipeline (bottom) applies guidance at selected steps, resulting in a shorter effective horizon ($m<n$), successful task completion (marked with \textcolor{green}{$\checkmark$}), and more robust execution.}

\label{fig:pipeline}
\end{figure*}

\subsection{Problem Formulation}
\label{sec:problem} 
VLA models aim to predict a sequence of executable actions $\boldsymbol{\mathcal{A}}=\{\mathbf{a}_{t+1:t+h}\}$ over a future prediction horizon of $h$ timesteps, by taking multi-modal information as input at timestep $t$, including visual observations $\mathbf{o}_{t}$, language instructions $\mathbf{z}$ (commonly represented as the text embedding vector), and robot states $\mathbf{s}_{t}=\{\mathbf{x}_{t},\bm{\theta}_{t}, grip_{t}\}$. This process can be formalized as:
\begin{gather}
    {\mathbf{a}}_{t+1:t+h} = {\mathbf{\pi_\theta}}(\mathbf{z}, {\mathbf{o}}_t, {\mathbf{s}}_t),
    \  
    \mathbf{a}_{i} = [\Delta \mathbf{x}, \Delta \bm{\theta}, \Delta grip]^{\top},
\end{gather}
where $\mathbf{\pi}_\theta$ represents the trained VLA model, and $\mathbf{a}_{t+1: t+h}$ denotes an action chunk with horizon size $h$. In the single-arm setting, each action $\mathbf{a}_i$ corresponds to a 7-DoF control vector, where $\Delta \mathbf{x} \in \mathbb{R}^3$ denotes the incremental Cartesian displacement of the end-effector relative to the previous step, $\Delta \bm{\theta} \in \mathbb{R}^3$ represents the incremental orientation offset in axis angles, and $\Delta grip \in \mathbb{R}$ specifies the change in the gripper command.
To achieve more realistic performance, we propose to enhance the visual input $\mathbf{o}_{t}$ during the inference process by incorporating both attention-guided and action-guided strategies.

We first define the sequential process to replace the original process at timestep $t$:
\begin{gather}
\mathbf{h}_t^{(\ell)} = \mathrm{Layer}_\ell\!\left(\pi_\theta;\, \mathbf{z}, \mathbf{o}_t, \mathbf{s}_t \right), 
\
\mathbf{a}_{t+1} = {\mathbf{\pi_\theta}}(\mathbf{z}, {\mathbf{o}}_t, {\mathbf{s}}_t)[0],  \\
\mathbf{M}^{\text{att}}_t = \Psi_{\text{att}}\!\left(\mathbf{h}_t^{(\ell)}\right), 
\
\mathbf{M}^{\text{act}}_t = \Psi_{\text{act}}\!\left(\mathbf{a}_{t+1}\right), \\
\mathbf{o}'_t = \mathcal{E}\!\Big(\mathbf{o}_t;\, 
\mathbf{M}_t\Big), \quad 
\mathbf{M}_t \in \big\{\mathbf{M}^{\text{att}}_t,\, \mathbf{M}^{\text{act}}_t\big\}, \\
\mathbf{a}_{t+1: t+h} = \pi_\theta\!\left(\mathbf{o}'_t,\mathbf{z},\mathbf{s}_t\right),
\label{eq:pipeline}
\end{gather}
where $\mathbf{h}_t^{(\ell)}$ denotes the hidden representation at the $\ell$-th attention layer, and $\mathbf{a}_{t+1}$ denotes the first action step of the predicted action chunk $\mathbf{a}_{t+1:t+h}$. $\Psi_{\text{att}}(\cdot)$ and $\Psi_{\text{act}}(\cdot)$ denote the mapping functions that generate the attention-guided mask $\mathbf{M}^{\text{att}}_t$ and the action-guided mask $\mathbf{M}^{\text{act}}_t$, respectively. The updated observation $\mathbf{o}'_t$ is obtained by applying the inference-time guidance operator $\mathcal{E}(\cdot)$ with either of the masks, \textit{i.e.,} $\mathbf{M}_t \in \{\mathbf{M}^{\text{att}}_t, \mathbf{M}^{\text{act}}_t\}$.

As illustrated in Fig.~\ref{fig:pipeline}, the original inference pipeline predicts actions solely from the raw observation $\mathbf{o}_t$. While this pipeline works reasonably well in many cases, it lacks the ability to leverage implicit reasoning available within the model. Consequently, an early misprediction at a single frame may propagate across the horizon, leading to cascading errors and eventual task failure. In contrast, our ATA framework introduces attention-guided and action-guided inference as implicit reasoning cues derived from attention maps and robot actions, which intervene directly on the observation stream at inference time. This implicit reasoning mechanism mitigates error propagation, improves robustness, and ultimately enhances the success rate of VLA models.
To address this issue, our framework introduces periodic inference-time guidance on visual observations through attention-guided and action-guided strategies. Rather than relying solely on the raw observation $\mathbf{o}_t$ at every step, we selectively replace $\mathbf{o}_t$ with the updated observation $\mathbf{o}'_t$ defined in Eq.~\eqref{eq:pipeline}. Concretely, we propose two complementary strategies:
\begin{itemize}
    \item \textbf{Attention-Guided Strategy:} An intermediate attention map from a designated layer of the VLA model is used as an implicit reasoning cue to highlight task-relevant objects and suppress irrelevant background regions, ensuring that the model’s perception remains aligned with the instruction.  
    \item \textbf{Action-Guided Strategy:} The robot’s end-effector state is leveraged to construct a directional region of interest (RoI), implicitly reasoning about the intended action trajectory by emphasizing regions along the motion direction while de-emphasizing unrelated areas.  
\end{itemize} 

By injecting the guided observations into the pipeline at selected frames, the model receives refined visual inputs that help maintain a correct reasoning trajectory. As illustrated in Fig.~\ref{fig:pipeline}, this stage-wise guidance enables the model to produce more reliable action predictions, thereby improving the overall task success rate. 
This naturally raises three key research questions:
\textit{1) How to design the attention-guided strategy?}
\textit{2) How to construct the action-guided strategy?}
\textit{3) How to adaptively apply these strategies during inference?}



\subsection{Attention-Guided}
\label{sec:attention}
To answer the first question, we begin by analyzing the structure of VLA blocks. Each block consists of Layer Normalization, an attention layer, another Layer Normalization, and an MLP, all connected through residual links to ensure stability and effective learning. The output token sequence of one block is then passed to the next. For the attention-guided strategy, we exploit the attention map generated inside the attention layer, since it reflects which regions of the input the model attends to during inference timestep $t$.
The input sequence includes both image and text tokens. Within the attention layer, each token computes similarity scores between its query vector and the key vectors of all other tokens, followed by a softmax normalization to produce attention weights. To construct our guidance signal, we focus on the last query token, which typically aggregates high-level contextual information across the entire sequence. From this token, we extract the attention weights corresponding only to the image tokens, and average them across all attention heads to obtain the aggregated attention map,  \(\Psi_{i,j}^{(L)} = \tfrac{1}{H} \sum_{h=1}^{H} Attn_h^{(L)}\), where \(Attn_h^{(L)}\) denotes the attention weights at layer \(L\) between the last query token and the image tokens for head \(h\). Here, $(i, j)$ indexes the spatial location of the image token in the patch grid, so that $\Psi_{i, j}^{(L)}$ represents the aggregated attention weight assigned to the patch at position $(i, j)$.

After extracting the attention map $\Psi_{i, j}^{(L)}$, we transform it into a mask that can serve as an inference-time guidance signal. To make the map effective, we first normalize it by subtracting the mean and dividing by the standard deviation across all pixels, followed by a sigmoid mapping:
\begin{gather}
    \mathbf{M}_t^{attn}(i, j) = \text{Sigmoid}\left( \frac{\Psi_{i,j}^{(L)} - \mu(\Psi)}{\sigma(\Psi)} \right),
    \label{eq:M_attn}
\end{gather}
Where $\mathbf{M}_t^{attn}(i, j)$ is the normalized attention mask, $\mu(\Psi)$ and $\sigma(\Psi)$ are the mean and standard deviation of the attention map $\Psi_{i,j}^{(L)}$, respectively. The sigmoid function maps the values to the range $[0, 1]$, ensuring that they are suitable for acting as a mask. Importantly, our attention maps are obtained by explicitly computing the similarity between queries and keys, independent of the value-update operation. This makes the procedure fully compatible with efficient implementations such as FlashAttention \cite{dao2022flashattention}, without affecting its acceleration benefits. The final step is to apply the enhanced attention map as a mask to the original image. This mask will highlight the most relevant regions of the image while suppressing irrelevant ones. The updated visual observation $\mathbf{o}_t^{'}$ is obtained as follows:
\begin{gather}
    \mathbf{o}_t' = \mathbf{o}_t \odot \mathbf{M}_t^{attn} + (1 - \mathbf{M}_t^{attn}) \odot \mathbf{bg}, 
    \label{eq:o_attn}
\end{gather}
where $\mathbf{M}_t^{attn}$ is broadcast to image channels, and $\mathbf{bg}$ is typically set to a neutral value like gray. This process yields guided observations that emphasize task-relevant areas while suppressing distractors, thereby providing implicit reasoning cues for the model during inference.

\subsection{Action-Guided}
\label{sec:action}
To answer the second question, we propose an action-guided strategy that exploits the robot’s end-effector (EEF) state as an implicit reasoning cue. Unlike the attention-guided strategy, which derives signals from the model’s internal representations, the action-guided strategy directly encodes the robot’s motion intent into a directional region of interest (RoI) on the image plane.

\begin{figure}[t]
  \includegraphics[width=1\linewidth]{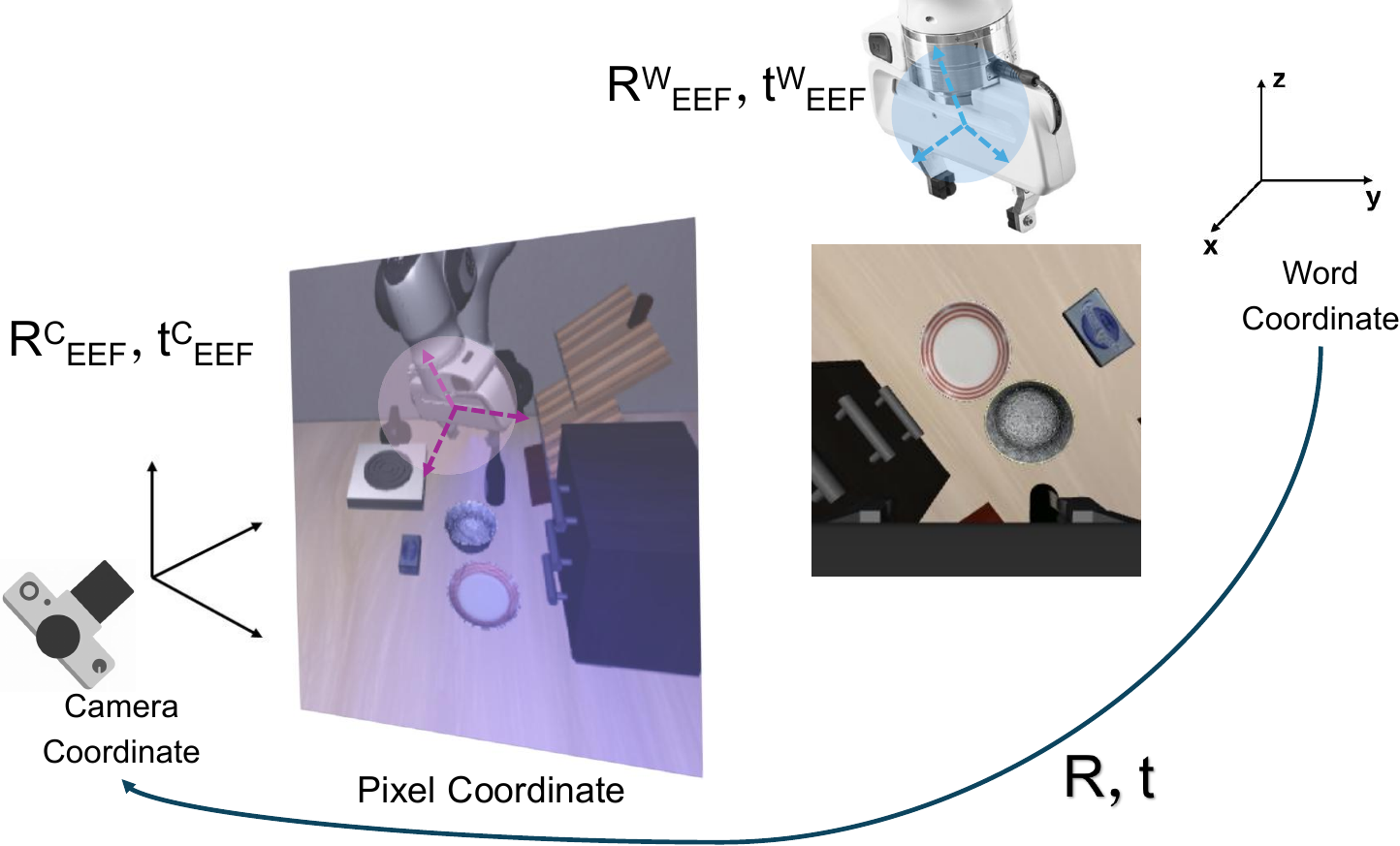}
  \caption{Illustration of the action-guided strategy. The end-effector (EEF) pose $(\mathbf{R}^{W}_{\text{EEF}},\mathbf{t}^{W}_{\text{EEF}})$ is obtained in the world coordinate system and transformed into the camera coordinate $(\mathbf{R}^{C}_{\text{EEF}},\mathbf{t}^{C}_{\text{EEF}})$, allowing projection of the EEF direction onto the image plane. The resulting region of interest (RoI) is represented as a conic sector in the pixel coordinate system. For visualization, we overlay a semi-transparent red mask so that pixels inside the RoI appear in red, while the actual inference mask is computed from Eq.~\eqref{eq:M_act}.
  }
  \label{fig:act}
  \vspace{-4mm}
\end{figure}

At timestep $t$, the EEF pose is given by $(\mathbf{x}_t,\bm{\theta}_t)$, where $\mathbf{x}_t \in \mathbb{R}^3$ is the Cartesian position and $\bm{\theta}_t$ is the axis angles converted from the orientation quaternion. From $\bm{\theta}_t$ we obtain the rotation matrix $\mathbf{R}_t$, and select a local tool axis (e.g., the $z$-axis) to define the intended motion direction $\mathbf{d}_t = \mathbf{R}_t \mathbf{e}_z$. Together with the base point $\mathbf{x}_t \in \mathbb{R}^3$, this direction specifies a conic sector parameterized by an opening angle $\alpha$ and a depth-adaptive length $z_{\text{depth}}$. 
To map this sector into the image, we use the camera parameters to project both the base point $\mathbf{x}_t$ and the tip point $\mathbf{x}_t + z_{\text{depth}} \mathbf{d}_t$ into the pixel plane, yielding the base pixel $(u_0,v_0)$ and the projected direction vector $(d_u,d_v)$. For each pixel $(u,v)$, we measure the angular deviation from this direction:

\begin{gather}
    \psi = 
    \frac{(u-u_0, v-v_0)\cdot(d_u,d_v)}
         {\|(u-u_0, v-v_0)\| \cdot \|(d_u,d_v)\|},
\end{gather}
where pixels with $\psi\geq \cos(\alpha/2)$ fall inside the sector. To provide smooth transitions at the boundary, we define a soft mask: 
\begin{gather}
    \mathbf{M}^{\text{act}}_{t}(u,v) =
    \max\!\left[
        \frac{\psi-\cos(\alpha/2)}{1-\cos(\alpha/2)}, \; 0
    \right].
    \label{eq:M_act}
\end{gather}

In practice, we set the opening angle to $\alpha=150^\circ$ (empirically, $\alpha \geq 120^\circ$ also works). This wide opening angle ensures that the manipulated object remains within the RoI while still emphasizing the intended motion direction. Moreover, a larger $\alpha$ produces smoother mask boundaries, which avoids jagged edges and maintains high visual quality during image blending.
Finally, the enhanced observation is obtained by applying this mask to the original image:
\begin{gather}
    \mathbf{o}'_t = \mathbf{o}_t \odot \mathbf{M}^{\text{act}}_{t} + (1-\mathbf{M}^{\text{act}}_{t})\odot \mathbf{bg},
    \label{eq:o_act}
\end{gather}
where $\mathbf{M}^{\text{act}}_{t}$ is broadcast to image channels, and $\mathbf{bg}$ is a neutral background (e.g., gray). As illustrated in Fig.~\ref{fig:act}, regions along the predicted motion direction are emphasized while irrelevant areas are suppressed, injecting action intent as an implicit reasoning signal into the inference pipeline.

\subsection{Inference-Time Integration}
\label{sec:inference}
To address the third question, we revisit the standard inference pipeline of VLA models. At each timestep $t$, the model predicts an action chunk from multimodal inputs. Since the initial robot state is usually fixed, variations in predictions depend largely on the quality of the visual observations. Early frames are especially critical: small errors at the beginning may propagate across the horizon and cause cascading failures. 

Our key insight is that injecting implicit reasoning cues into the observation stream early in the task can stabilize the reasoning trajectory. Specifically, we apply \textbf{attention-guided reasoning} at the very first frame to direct the model’s focus toward task-relevant regions, thereby improving the prediction of the initial action chunk. This corresponds to the inference pipeline in Eq.~\eqref{eq:M_attn} with $\mathbf{M}_t^{attn}$.
While attention-guided reasoning improves semantic understanding of the scene, it does not fully leverage the robot’s physical interaction state. To complement it, we apply \textbf{action-guided reasoning} based on the EEF pose during the early stages of the task. This strategy highlights regions along the intended motion direction, ensuring that the model perceives both the semantic goal and the geometric action intent. Formally, this corresponds to Eq.~\eqref{eq:M_act} with $\mathbf{M}_t^{act}$.

In practice, we integrate the proposed two strategies by applying attention-guided reasoning at the first frame and action-guided reasoning in the early steps. 
This combined schedule improves action predictions with minimal computational overhead, as it only introduces an extra forward pass when guidance is applied. 
We find that periodic application of attention-guided reasoning cues further boost performance; however, overly frequent usage may inject noise into the visual stream and reduce efficiency, as discussed in Section~\ref{sec:exp}.

\subsection{Overall Algorithm}
\label{sec:alg}
The sequential inference process of ATA is summarized in Alg.~\ref{alg:1}, where inference ends upon task success or when the maximum step $M$ is reached. 
At each timestep $t$, the model receives the visual observation $\mathbf{o}_t$, instruction $\mathbf{z}$, and state $\mathbf{s}_t$, and outputs an action chunk. 
To inject implicit reasoning, we apply two guidance functions: $f^{attn}$ (Sec.~\ref{sec:attention}) and $f^{act}$ (Sec.~\ref{sec:action}). These functions generate masks that update the observation before it enters the policy $\pi_\theta$. Here $bg$ is a neutral background, $Freq$ is the trigger frequency of Attention-Guided reasoning, and $i^{act}$ is the timestep for Action-Guided reasoning.

\renewcommand{\algorithmicrequire}{\textbf{Inputs:}}
\renewcommand{\algorithmicensure}{\textbf{Outputs:}}
\begin{algorithm}[t]
\caption{ATA Inference Algorithm}\label{alg:1}
\KwIn{$\mathbf{o}, \mathbf{z}, \mathbf{s}, \pi_{\theta}, M, Freq, i^{act}, bg, f^{attn}, f^{act}$}
\KwOut{Action chunk $\mathcal{A}$}

$i \gets 0, \; \mathbf{s} \gets \mathbf{s}_0$\;
\Repeat{success or $i \geq M$}{
    \If{$i \bmod Freq = 0$}{
        $\mathbf{M}_i^{attn} \gets f^{attn}(\pi_{\theta}; \mathbf{z}, \mathbf{o}_i, \mathbf{s}_i)$ \Comment{Eq.~\ref{eq:M_attn}}
        $\mathbf{o}_i \gets \mathbf{o}_i \odot \mathbf{M}_i^{attn} + (1-\mathbf{M}_i^{attn}) \odot bg$ \Comment{Eq.~\ref{eq:o_attn}}
    }
    \If{$i = i^{act}$}{
        $\mathbf{M}_i^{act} \gets f^{act}(\pi_{\theta}; \mathbf{z}, \mathbf{o}_i, \mathbf{s}_i)$ \Comment{Eq.~\ref{eq:M_act}}
        $\mathbf{o}_i \gets \mathbf{o}_i \odot \mathbf{M}_i^{act} + (1-\mathbf{M}_i^{act}) \odot bg$ \Comment{Eq.~\ref{eq:o_act}}
    }
    $\mathcal{A}_{i+1} \gets \pi_{\theta}(\mathbf{z}, \mathbf{o}_i, \mathbf{s}_i)$\;
    $i \gets i + 1$\;
}
\end{algorithm}

 \section{Experiments}
\label{sec:exp}

\subsection{Experimental Settings}
\subsubsection{Model Settings} 
To demonstrate the effectiveness of our proposed method, we apply it to several state-of-the-art VLA models, including OpenVLA, $\pi_{0}$-fast, HybridVLA, and GR00T-N1.5 models, and evaluate them across both simulation and real-world environments. Specifically, we first evaluate the performance of OpenVLA and $\pi_0$-fast models in the LIBERO simulation enviornment, and the HybridVLA model in the RLBench simulation enviornment. Additionally, we perform real-world experiments using the GR00T-N1.5 model to validate the practical effectiveness of ATA.

\begin{table}[t]
\centering
\setlength{\tabcolsep}{8pt}
\caption{\textbf{Best configurations of Attention-Guided and Action-Guided strategies.} 
\textit{L} denotes the attention layer index, \textit{Freq} denotes the trigger frequency.
For $\pi_{0}$-fast and HybridVLA, the attention-guided strategy is applied only once ($\textit{Freq} = \infty$) at the first timestep ($t = 0$).}
\scalebox{0.88}{
\begin{tabular}{lccccc}
\toprule
Model & Dataset & L & Freq & Time & Step\\
\midrule
\multirow{4}{*}{OpenVLA} & Spatial & 20 & 100 & periodic & 50\\
        & Goal    & 5  & 100 & periodic & 80\\
        & Object  & 20 & 100 & periodic & 60 \\
        & 10      & 20 &  50 & periodic & 170 \\
\midrule
$\pi_{0}$-fast & LIBERO       & 16/12/12/12 & $\infty$ & $t=0$ & 10/7/5/2 \\
HybridVLA      & RLBench      & 5           & $\infty$ & $t=0$ & 3\\
\bottomrule
\end{tabular}
}
\label{tab:config}
\vspace{-2mm}
\end{table}

\subsubsection{Evaluation and Datasets} 
In the LIBERO simulation environment, we evaluate our method on four datasets: \textit{LIBERO-Spatial}, \textit{LIBERO-Goal}, \textit{LIBERO-Object}, and \textit{LIBERO-Long}, each containing $500$ demonstrations across $10$ tasks. Specifically, LIBERO-Spatial focuses on spatial generalization with identical objects placed in different spatial relations, LIBERO-Object evaluates generalization to novel objects with a unique object in each task, LIBERO-Goal tests goal generalization with fixed objects and layouts but varying task goals, and LIBERO-Long consists of long-horizon tasks to assess the model’s ability to handle extended action sequences. 
In the RLBench environment, we test the model on $8$ tasks\footnote{Tasks include: close box, toilet seat down, sweep to dustpan, close fridge, close laptop, take umbrella out, wine at rack, water plants}. 

In the real world, we collect a dataset of $500$ stacking trajectories in a $90cm\times50cm$ workspace, where blue and red blocks are randomly placed at predefined positions. Each trajectory, built with $3cm\times3cm\times3cm$ blocks, consists of robot actions, language instruction, and visual observations from three cameras (head, wrist, and external). The dataset includes $100$ one-layer, $150$ two-layer, and $250$ three-layer towers, covering both simple and challenging tasks while balancing diversity and collection cost for fine-tuning GR00T-N1.5. For evaluation, we randomly sample $50$ distinct block configurations and locations to compare performance with and without ATA. The real-world robotic system is shown in the left part of Tab.~\ref{tab:rw}, featuring a $7$-DoF Discover ARM with wrist, head, and external cameras.

\begin{table*}[t]
    \centering
    \setlength{\tabcolsep}{8pt}
    \caption{\textbf{Comparison of ATA and Baselines on LIBERO and RLBench.} LIBERO includes four datasets, each with 10 tasks and 50 demonstrations per task. The RLBench environment evaluates 8 tasks, each with 100 demonstrations. \textit{Avg S.R.} denotes the average success rate. \textit{Attn} denotes the use of only the Attention-Guided strategy, and \textit{ATA} integrates both Attention-Guided and Action-Guided strategies. \textit{Avg S.I.C.} indicates the average number of inference calls required for successfully completed tasks, and \textit{Avg I.C.} represents the overall average number of inference calls across all tasks.} 
    \resizebox{\textwidth}{!}{
        \begin{tabular}{ccccccccccccc}
            \toprule
            Models                  & Method                  & \multicolumn{2}{c}{LIBERO-Spatial} & \multicolumn{2}{c}{LIBERO-Goal} & \multicolumn{2}{c}{LIBERO-Object} & \multicolumn{2}{c}{LIBERO-Long} & Avg S.R. & Avg S.I.C$\downarrow$ & Avg I.C.$\downarrow$                      \\
            \toprule
            OpenVLA                 & Baseline                & \multicolumn{2}{c}{84.2}           & \multicolumn{2}{c}{75.6}        & \multicolumn{2}{c}{88.6}          & \multicolumn{2}{c}{55.2}        & 75.9     & 165   & 235                         \\
            OpenVLA                 & API                     & \multicolumn{2}{c}{86.2}           & \multicolumn{2}{c}{78.2}        & \multicolumn{2}{c}{88.6}          & \multicolumn{2}{c}{52}          & 76.3     & 167   & 232  
                                   \\ 
            OpenVLA                 & ATTN                    & \multicolumn{2}{c}{87}             & \multicolumn{2}{c}{82}          & \multicolumn{2}{c}{92}            & \multicolumn{2}{c}{57.4}        & 79.6     & 165   & 228                            \\
            \rowcolor{gray!20}
            OpenVLA                 & ATA                     & \multicolumn{2}{c}{88.8}           & \multicolumn{2}{c}{82.6}        & \multicolumn{2}{c}{94.6}          & \multicolumn{2}{c}{58.2}        & 81.1     & 164   & 225                            \\
            \midrule
            $\pi_0$-fast             & Baseline                & \multicolumn{2}{c}{96.2}           & \multicolumn{2}{c}{89}          & \multicolumn{2}{c}{98.2}          & \multicolumn{2}{c}{60.2}        & 85.9     & 30   & 41                            \\
            $\pi_0$-fast             & API                & \multicolumn{2}{c}{97.6}           & \multicolumn{2}{c}{88.8}          & \multicolumn{2}{c}{98}          & \multicolumn{2}{c}{60.8}           & 86.3     & 31    & 41                            \\            
            $\pi_0$-fast             & ATTN                    & \multicolumn{2}{c}{98.2}           & \multicolumn{2}{c}{90.2}        & \multicolumn{2}{c}{98.6}          & \multicolumn{2}{c}{62.6}        & 87.3     & 30    & 40                            \\
            \rowcolor{gray!20}
            $\pi_0$-fast             & ATA                     & \multicolumn{2}{c}{98.6}           & \multicolumn{2}{c}{90.8}        & \multicolumn{2}{c}{98.8}          & \multicolumn{2}{c}{63.2}        & 87.9     & 29     & 39                            \\
            \midrule
            \midrule
            \multirow{2}{*}{Models} & \multirow{2}{*}{Method} & \multirow{2}{*}{\makecell{Close                                                                                                                                                      \\ box}} & \multirow{2}{*}{\makecell{Close \\ laptop lid}} & \multirow{2}{*}{\makecell{Toilet \\ seat down}} & \multirow{2}{*}{\makecell{Sweep \\ to dustpan}} & \multirow{2}{*}{\makecell{Close \\ fridge}} & \multirow{2}{*}{\makecell{Umbrella \\ out}} & \multirow{2}{*}{\makecell{Wine at \\ rack}} & \multirow{2}{*}{\makecell{Water \\ plants}} & \multirow{2}{*}{\makecell{Avg S.R.}} & {\multirow{2}{*}{Avg S.I.C.$\downarrow$}} & {\multirow{2}{*}{Avg I.C.$\downarrow$}} \\
            \\
            \midrule
            HyBridVLA               & Baseline                & 85                                 & 90                              & 90                                & 100                             & 90       & 40      & 50 & 25 & 71.3 & 7 & 8 \\
            HyBridVLA               & ATTN                    & 87                                 & 92                              & 92                                & 97                              & 95       & 43      & 53 & 30 & 73.6 & 7 & 7 \\
            \rowcolor{gray!20}
            HyBridVLA               & ATA                     & 88                                 & 95                              & 100                               & 98                              & 97       & 46      & 55 & 35 & 76.8 & 6 & 7 \\
            \bottomrule
        \end{tabular}
    }
    \label{tab:sim}
\end{table*}

\begin{table*}[t]
    \centering
    \setlength{\tabcolsep}{8pt}
    \caption{\textbf{Comparison of Blur, RandomBlur, Attention-Guided, and Baselines on LIBERO and RLBench datasets.} \textit{Blur} represents the Gaussian Blur applied to the first frame, \textit{RandomBlur} denotes the Gaussian Blur applied randomly to frames after the first frame, and \textit{ATTN*} indicates the attention-guided strategy applied only on the first frame.}
    \resizebox{\textwidth}{!}{
        \begin{tabular}{ccccccccccc}
            \toprule
            Models                  & Method                  & \multicolumn{2}{c}{LIBERO-Spatial} & \multicolumn{2}{c}{LIBERO-Goal} & \multicolumn{2}{c}{LIBERO-Object} & \multicolumn{2}{c}{LIBERO-Long} & Avg S.R.                       \\
            \toprule
            OpenVLA                 & Baseline                & \multicolumn{2}{c}{84.2}           & \multicolumn{2}{c}{75.6}        & \multicolumn{2}{c}{88.6}          & \multicolumn{2}{c}{55.2}        & 75.9                           \\
            OpenVLA                 & Blur                    & \multicolumn{2}{c}{83.4}           & \multicolumn{2}{c}{74.6}        & \multicolumn{2}{c}{84.2}          & \multicolumn{2}{c}{53.2}        & 73.9                           \\
            OpenVLA                 & RandomBlur              & \multicolumn{2}{c}{84.6}           & \multicolumn{2}{c}{77.2}        & \multicolumn{2}{c}{87.2}          & \multicolumn{2}{c}{53.8}        & 75.7                           \\
            \rowcolor{gray!20}
            OpenVLA                 & ATTN*                   & \multicolumn{2}{c}{86.6}           & \multicolumn{2}{c}{80.8}        & \multicolumn{2}{c}{89.2}          & \multicolumn{2}{c}{56.4}        & 78.3                           \\
            \midrule
            \midrule
            \multirow{2}{*}{Models} & \multirow{2}{*}{Method} & \multirow{2}{*}{\makecell{Close                                                                                                                                             \\ box}} & \multirow{2}{*}{\makecell{Close \\ laptop lid}} & \multirow{2}{*}{\makecell{Toilet \\ seat down}} & \multirow{2}{*}{\makecell{Sweep \\ to dustpan}} & \multirow{2}{*}{\makecell{Close \\ fridge}} & \multirow{2}{*}{\makecell{Umbrella \\ out}} & \multirow{2}{*}{\makecell{Wine at \\ rack}} & \multirow{2}{*}{\makecell{Water \\ plants}} & \multirow{2}{*}{\makecell{Avg S.R.}} \\
            \\
            \midrule
            HyBridVLA               & Baseline                & 85                                 & 90                              & 90                                & 100                             & 90       & 40 & 50 & 25 & 71.3 \\
            HyBridVLA               & Blur                    & 40                                 & 45                              & 25                                & 20                              & 85       & 35 & 15 & 5  & 42.7 \\
            HyBridVLA               & RandomBlur              & 55                                 & 85                              & 95                                & 20                              & 75       & 50 & 20 & 25 & 53.1 \\
            \rowcolor{gray!20}
            HyBridVLA               & ATTN*                   & 87                                 & 92                              & 92                                & 97                              & 95       & 43 & 53 & 30 & 73.6 \\
            \bottomrule
        \end{tabular}
    }
    \label{tab:ab1}
\end{table*}

\subsubsection{Implementation Details} 
For the OpenVLA and $\pi_{0}$-fast experiments in the LIBERO environment, as well as the HybridVLA experiments in the RLBench environment, all experiments are conducted on a single H20 GPU. For the GR00T-N1.5 experiments, we perform training on $8$ NVIDIA A$800$ GPUs with mixed-precision training, resulting in a total training time of $35$ hours over $90,000$ steps. 

In our experiments, we explore different configurations of the Attention-Guided and Action-Guided strategies. As shown in Tab.~\ref{tab:config}, for OpenVLA, the best results are obtained with dataset-specific layer choices $L \in \{20,5,20,20\}$ and periodic triggers $Freq \in  \{100,100,100,50\}$ across LIBERO-Spatial, Goal, Object, and Long, respectively. For $\pi_{0}$-fast, due to its long-horizon action chunking, the Attention-Guided strategy is more effective when applied only once at the first timestep ($t=0$). Similarly, for HybridVLA in RLBench, where tasks require only short action sequences, Attention-Guided reasoning also achieves the best performance when triggered once at $t=0$. For all models, the Action-Guided strategy is applied only at the initial timestep, since early intervention provides the most benefit for guiding subsequent actions.

To ensure fairness, both LIBERO and RLBench experiments are executed under maximum step limits, where exceeding the predefined horizon is considered a task failure. For LIBERO, we follow the configurations in OpenVLA and $\pi_{0}$, with maximum step lengths of $220$ for Spatial, $300$ for Goal, $280$ for Object, and $520$ for LIBERO-10. For RLBench, consistent with HybridVLA, we set the maximum horizon to $10$ steps for all tasks.

\subsection{Main Results}
\subsubsection{Simulation Results} The results presented in Tab.~\ref{tab:sim} show the performance of the models after applying our ATA framework, with the \textit{Average Success Rates} (S.R.), \textit{Average Number of Inference Calls} for successfully completed tasks (S.I.C), and \textit{Overall Average Number of Inference Calls} (I.C.). 
For the API \cite{yu2024attention} method, the improvements are relatively limited since it only leverages the ViT component and does not incorporate deeper attention or action information. Specifically, on the LIBERO dataset, API brings only a $0.4\%$ gain for OpenVLA and $\pi_0$-fast, while in RLBench, it improves HybridVLA by just $1.7\%$ over the baseline. When only the Attention-Guided method is used, compared to the baseline, OpenVLA and $\pi_0$-fast models show improvements of $3.7\%$ and $1.4\%$ respectively on the LIBERO dataset. When combined with the Action-Guided method, the performance improves further to $5.2\%$ for OpenVLA and $2\%$ for $\pi_0$-fast, highlighting the complementary nature of the two strategies. Similarly, for the HybridVLA model tested in the RLBench, the Attention-Guided method alone provides a $2.3\%$ improvement over the baseline. When the Action-Guided method is integrated, the improvement increases to $5.5\%$, demonstrating the additive benefits of combining both strategies.

Notably, ATA improves both performance and efficiency. For OpenVLA, the average inference calls drop from $235$ to $225$ (Avg I.C.) and from $165$ to $164$ (Avg S.I.C.). Similar reductions are observed for $\pi_0$-fast ($41\rightarrow39$, $30\rightarrow29$) and HybridVLA ($8\rightarrow7$, $7\rightarrow6$). Although ATA requires an extra step to introduce implicit reasoning, it ultimately lowers the overall number of inference calls. Since simulator runs involve costly environment resets, we report average inference calls instead of wall-clock time as a reliable measure of efficiency.

\begin{table}[t]
    \centering
    \setlength{\tabcolsep}{8pt}
    \caption{\textbf{Comparison of Different Frequencies of Attention-Guided Strategy on LIBERO Dataset Performance.} \textit{0} denotes the baseline, \textit{$\infty$} denotes applying Attention-Guided (\textit{ATTN*}) only on the first timestep (\textit{$t=0$}), and \textit{20}, \textit{50}, \textit{100}, \textit{200} indicate applying Attention-Guided every 20, 50, 100, and 200 inference steps, respectively.}
    \scalebox{0.95}{
        \begin{tabular}{cccccc}
            \toprule
            Models  & Freq             & Spatial        & Goal & Object & Long \\
            \toprule
            OpenVLA & 0 (Baseline)     & 84.2           & 75.6 & 88.6   & 55.2 \\
            OpenVLA & $\infty$ (ATTN*) & 86.6           & 80.8 & 89.2   & 56.4 \\
            OpenVLA & 20               & 83.9           & 78.8 & 87.8   & 55.6 \\
            OpenVLA & 50               & 86             & 79.6 & 89.2   & \underline{57.4} \\
            OpenVLA & 100              & \underline{87} & \underline{82}   & \underline{92}     & 56.4 \\
            OpenVLA & 200              & 86.2           & 81.2 & 89.6   & 55.8 \\
            \bottomrule
        \end{tabular}
    }
    \label{tab:ab2}
\end{table}

\subsubsection{Real-World Results} In the real world, we collected $500$ stacking block trajectories, where blocks are stacked between one, two, and three towers with red and blue colors. We validated the performance using a single-arm $7$-DoF Discover Arm. As shown in Tab.~\ref{tab:rw}, the ATA method improves the performance by $2\%$, $2\%$, and $6\%$ for stacking one, two, and three towers of blocks, respectively, compared to the baseline. Additionally, during the three-tower stacking test, we introduced irrelevant and unseen objects such as blocks of different colors, pens, scissors, etc. We found that the ATA method further improves robustness, achieving a $10\%$ increase in performance compared to the baseline.

The bottom part of Tab.~\ref{tab:rw} illustrates task execution sequences: the top sequence depicts stacking a three-layer tower in a simple scene containing only red blocks, corresponding to the language instruction ``\textit{Pick up three red blocks and stack them on the red point to form a three-layer tower.}''. In contrast, the bottom sequence depicts stacking a three-layer tower in a more complex scene with unseen distractor objects such as scissors, pens, and additional colored blocks (orange and green), guided by the instruction ``\textit{Pick up three blue blocks and stack them on the red point to form a three-layer tower.}''.

\subsection{Ablation Studies.}
\subsubsection{The Impact of the First Frame on Success Rates}
As discussed in Sec.~\ref{sec:inference}, the first frame plays a crucial role in setting the task context. To validate this, we compare three scenarios on OpenVLA (LIBERO) and HybridVLA (RLBench): \textit{1)} Gaussian Blur ($9\times9$) on the first frame; \textit{2)} Gaussian Blur on random subsequent frames; \textit{3)} Attention-Guided strategy on the first frame.

As shown in Tab.~\ref{tab:ab1}, blurring the first frame causes a $2\%$ drop on LIBERO and a $28.6\%$ drop on RLBench, whereas blurring random later frames only leads to $0.2\%$ and $18.2\%$ drops, respectively. In contrast, applying the Attention-Guided strategy to the first frame improves performance by $2.4\%$ on LIBERO and $2.3\%$ on RLBench. The larger impact on RLBench is due to its shorter task horizons, which amplify early frame effects.

\begin{table}[t]
    \centering
    \setlength{\tabcolsep}{8pt}
    \caption{\textbf{Real-world stacking results with and without ATA.} Performance is measured on stacking \textit{one}, \textit{two}, and \textit{three}-layer towers of blocks. For the three-tower case, we also tested robustness by adding irrelevant and unseen objects (\textit{Three*}). The bottom-left shows the real-world robotic setup, while the right part illustrates task execution sequences in both simple (\textit{top}) and complex (\textit{bottom}) scenes.}
    \scalebox{0.93}{
        \begin{tabular}{cccccc}
            \toprule
            Models  & Method            & One         & Two     & Three & Three* \\ 
            \toprule
            Gr00t-N1.5 & Baseline       & 92                & 85            & 68  & 46 \\
            Gr00t-N1.5 & ATA            & 94                & 87            & 74  & 56 \\      
            \bottomrule
            
            \multicolumn{6}{c}{
                \includegraphics[width=1\linewidth]{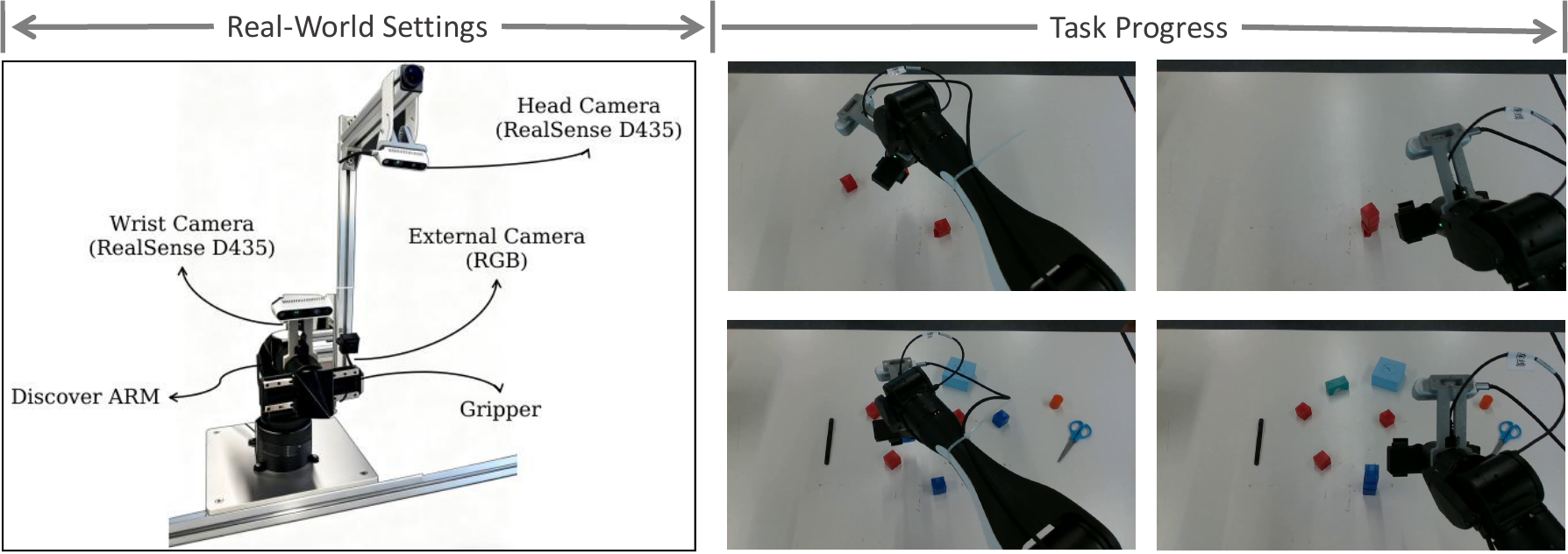}
            }
            
        \end{tabular}
    }
    \label{tab:rw}
\end{table}


\subsubsection{The Impact of Different Frequencies of Attention-Guided Strategy} 
As shown in Tab.~\ref{tab:ab2}, applying the Attention-Guided strategy only at the first timestep ($t=0$, \textit{ATTN*}) improves the \emph{average} success rate by $2.4\%$ over the baseline. We then vary the trigger frequency on LIBERO. When the frequency is set between $50$ and $100$ inference steps, the model attains the best \emph{average} performance, yielding a $3.7\%$ gain over the baseline and a $1.3\%$ gain over \textit{ATTN*}. In contrast, excessively sparse (e.g., every $200$ steps) or overly frequent (e.g., every $20$ steps) triggers degrade performance, indicating diminishing returns.

\section{Conclusion}
\label{sec:conclusion}

In this work, we presented ATA, a training-free framework that introduces implicit reasoning for VLA models through attention-guided and action-guided strategies. ATA adaptively refines visual inputs, remains compatible with efficient attention implementations, and requires no additional annotations or retraining. Experiments show that ATA improves task success and robustness while preserving, and even enhancing, inference efficiency, offering a lightweight yet effective path toward scalable VLA deployment. Looking forward, ATA provides a general paradigm for incorporating reasoning into VLA models without costly data collection, and can be extended to broader multimodal reasoning and control tasks. As future work, we aim to develop adaptive mechanisms for automatically selecting the optimal attention layer and trigger frequency in the Attention-Guided strategy, as well as determining the timing of Action-Guided intervention, thereby reducing dataset-specific tuning and further improving the generality of ATA.

\bibliographystyle{IEEEtran}
\bibliography{reference}

\end{document}